\documentclass[conference]{IEEEtran}
\IEEEoverridecommandlockouts
\usepackage{cite}
\usepackage{amsmath,amssymb,amsfonts}
\usepackage{algorithmic}
\usepackage{graphicx}
\usepackage{textcomp}
\usepackage{xcolor}
\def\BibTeX{{\rm B\kern-.05em{\sc i\kern-.025em b}\kern-.08em
    T\kern-.1667em\lower.7ex\hbox{E}\kern-.125emX}}
\begin{document}

\title{Evaluating the trustworthiness of the Fr\'{e}chet Inception Distance with stochastic embedding representations\\
\footnotesize\thanks{The project 22HLT05 MAIBAI has received funding from the European Partnership on Metrology, co-financed from
the European Union’s Horizon Europe Research and Innovation Programme and by the Participating States. Funding
for the UK partners was provided by Innovate UK under the Horizon Europe Guarantee Extension}}

\author{\IEEEauthorblockN{Ciaran Bench}
\IEEEauthorblockA{\textit{Department of Data Science and AI} \\
\textit{National Physical Laboratory}\\
Teddington, UK \\
ciaran.bench@npl.co.uk}
\and
\IEEEauthorblockN{Vivek Desai}
\IEEEauthorblockA{\textit{Department of Data Science and AI} \\
\textit{National Physical Laboratory}\\
Teddington, UK \\
vivek.desai@npl.co.uk}
\and
\IEEEauthorblockN{Carlijn Roozemond}
\IEEEauthorblockA{\textit{Dutch Expert Centre for Screening (LRCB)} \\
Nijmegen, The Netherlands \\
c.roozemond@lrcb.nl}
\and
\IEEEauthorblockN{Ruben van Engen}
\IEEEauthorblockA{\textit{Dutch Expert Centre for Screening (LRCB)} \\
Nijmegen, The Netherlands \\
r.vanengen@lrcb.nl}
\and
\IEEEauthorblockN{Spencer A. Thomas}
\IEEEauthorblockA{\textit{Department of Data Science and AI} \\
\textit{National Physical Laboratory}\\
Teddington, UK \\
spencer.thomas@npl.co.uk}
}

\maketitle

\begin{abstract}

Feature embeddings acquired from pretrained models are widely used in medical applications of deep learning to assess the characteristics of datasets; e.g. to determine the quality of synthetic, generated medical images. The Fr\'{e}chet Inception Distance (FID) is one popular synthetic image quality metric that relies on the assumption that the characteristic features of the data can be detected and encoded by an InceptionV3 model pretrained on ImageNet1K (natural images). While it is widely known that this makes it less effective for applications involving medical images, the extent to which the metric fails to capture meaningful differences in image characteristics is not obviously known. Here, we use Monte Carlo dropout to compute the predictive variance in the FID as well as a supplemental estimate of the predictive variance in the feature embedding model's latent representations. We show that the magnitudes of the predictive variances considered exhibit varying degrees of correlation with the extent to which test inputs (ImageNet1K validation set augmented at various strengths, and other external datasets) are out-of-distribution relative to its training data, providing some insight into the effectiveness of their use as indicators of the trustworthiness of the FID. 
\end{abstract}


\section{Introduction}
\label{sec:intro}

Deep generative models are capable of generating high quality synthetic medical images for a range of tasks: e.g. data augmentation \cite{kebaili2023deep} and image enhancement \cite{iglesias2022accurate}. For the generic task of random image generation, the aim is to effectively learn to sample from the target data distribution $p(x)$ given a discrete set of samples $X$ of $N$ images; $X = x_1, x_2, ..., x_N$. High quality generated images $\hat{x}$ should appear as though they have been sampled from this ground truth target distribution. However $p(x)$ is typically not known, and so many image quality metrics aim to compare the similarity between $X$ and a set of generated images $\hat{X} = \hat{x_1}, \hat{x_2}, ..., \hat{x_N}$.

\subsection{Assessing synthetic image quality with network embeddings}

One strategy to assess synthetic image quality is to compare the feature representations output by an intermediate layer of a deep learning model (i.e. a source model $G_s$) trained on a source task $\tau_s$ where large amounts of data (source data $D_s(X_s,Y_s)$, where $x_1, x_2, ..., x_N \in X_s$ and $y_1, y_2, ..., y_N \in Y_s$) are available. However, the success of this approach is contingent on $G_s$'s ability to extract/encode features $f_t$ relevant to the target task $\tau_t$ from the target data $D_t$. Broadly, this requires that:

\begin{itemize}
    \item $f_t$ are represented in $D_s$,
    \item $f_t$ are relevant to performing $\tau_s$ so that $G_s$ is in principle capable of extracting them from data,
    \item $G_s$ can generalise its ability to extract and encode $f_t$ from $D_t$ despite their possible discrepancy in representation compared to $D_s$, 
    \item $f_t$ is represented in the embedding in a manner that that is useful for performing/learning $\tau_t$ by whatever means are chosen.
\end{itemize}

\subsubsection{The Fr\'{e}chet Inception Distance (FID)}
In the context of synthetic image quality assessment, $\tau_t$ involves comparing the similarity in the semantic content of a generated set of images $\hat{X_t}$ with an unpaired reference set of real images $X_t$. We consider one popular metric that performs this task: the Fr\'{e}chet Inception Distance (FID) \cite{heusel2017gans} that compares stylistic content by assessing the difference in the distributions drawn over the activations produced by an InceptionV3 model pretrained on ImageNet1K (ILSVRC 2012) \cite{ILSVRC15}, when applied to $X_t$ and $\hat{X_t}$. Specifically, the mean $\mu_{x}$ and covariance $\Sigma_{x}$ of the activation maps produced by processing all the images in $X_t$ using the pretrained InceptionV3 model are used to parameterise a multivariate Gaussian distribution $\mathcal{N}_{{x}}(\mu_{x}, \Sigma_{x})$, that is then compared to the corresponding Gaussian computed from $\hat{X}$ $\mathcal{N}_{\hat{x}}(\mu_{\hat{x}}, \Sigma_{\hat{x}})$ with a Fr\'{e}chet distance:
\begin{equation}   
\begin{split}
        \mathrm{FID}(\hat{X},X) = d_F\left(\mathcal{N}_{{\hat{x}}}(\mu_{\hat{x}}, \Sigma_{\hat{x}}), \mathcal{N}_{x}\left(\mu_x, \Sigma_x\right)\right)^2= \\
        \left\|\mu_{\hat{x}}-\mu_x\right\|_2^2+\text{tr}\left(\Sigma_{\hat{x}}+\Sigma_x-2\left(\Sigma_{\hat{x}} \Sigma_x\right)^{\frac{1}{2}}\right).
    \end{split}
\end{equation}

In our case study, ImageNet1K is $D_s$, and the pretrained InceptionV3 is $G_s$. The suitability of $G_s$ depends on its ability to encode semantically relevant information from $\hat{X}$ and $X$ in a manner that lends itself to the calculation of the FID. This second criterion is satisfied by our use of the InceptionV3 for $G_s$. 

It is widely  known that FID is less suitable for medical image datasets (among other types of non-natural images) given they are out of distribution relative to the training set of the feature embedding model \cite{konz2024fr,wu2025pragmatic,truong2021transferable,kynkaanniemi2022role}. I.e. the feature embedding model is likely to provide less informative feature representations for medical images. 

Uncertainty in measurement outputs support decision making and provide confidence in measured values. In this work we investigate whether the use of uncertainty quantification (UQ) techniques can help assess the quality of the embedding representations, and aid in the interpretation of how effective/informative the FID may be on non-natural datasets.

\subsection{Aim: evaluating the trustworthiness of the FID with MCD}
An estimate of the doubt in the feature embedding model's ability to encode characteristic feature representations from images could be used to help interpret the effectiveness/trustworthiness of the FID, or analogous metrics that utilise a different feature embedding model (e.g. the Fr\'{e}chet Autoencoder distance \cite{buzuti2023frechet,bench2025quantifying}). UQ techniques provide a means to acquire the predictive variance of embedding representations that may be associated with the uncertainty in the embedding representation resulting from variations in input data quality or any discrepancy between the training and test data domains \cite{nguyenstochastic,an2023maximum,ahn2024uncertainty,scott2019stochastic,oh2018modeling}. Bayesian optimisation is a popular method with which one may derive epistemic uncertainty from a predictive model. While in principle, this may be applied to neural networks (Bayesian Neural Networks \cite{jospin2022hands}), in practice, the technique does not scale well with the number of parameters involved. Approximation schemes like variational inference also fail to scale to larger models, spurring the development of even more efficient approaches. Here, we employ Monte Carlo dropout (MCD) \cite{gal2016dropout}, which approximates variational inference by training and evaluating inputs with dropout regularisation applied liberally throughout the architecture. This approach is computationally efficient, straightforward to implement, stable, and avoids issues with reparameterisation associated with Deep ensembles (another scalable UQ technique that models epistemic uncertainty). We examine whether the predictive variance of embedding representations estimated with MCD, and the corresponding predictive variance in the FID, may provide a heuristic indication of whether the model encodes characteristic feature representations of the data (given its reported sensitivity to out of distribution (OOD) inputs and data quality), and therefore, the effectiveness of the FID. 

\subsection{Behaviour of MCD on various datasets}
We employ data augmentation as a tool to observe the behaviour of the predictive variance in the embeddings/FID as the properties of input data vary. We use the resulting observations to supplement discussion about how well the predictive variances reflect the trustworthiness of the FID.

\begin{figure*}[h!]
    \centering\includegraphics[width=1.99\columnwidth]{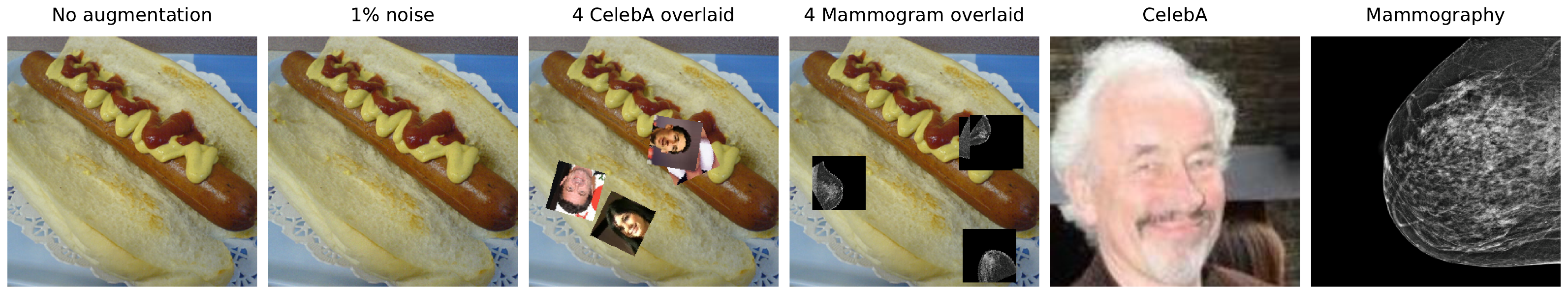}

    \caption{Example inputs from left to right: an ImageNet1K validation set image, the same image augmented with additive random Gaussian noise with a standard deviation defined by 1~$\%$ of the maximum amplitude, augmented with four random miniaturised CelebA images randomly rotated and overlaid, augmented with four random miniaturised mammography images randomly rotated and overlaid, a random CelebA image, and a mammogram.}
    \label{fig:augs}
\end{figure*}

We also compute the predictive variance of the embeddings/FID for an \textit{unaugmented} reference set and various test datasets that are increasingly OOD relative to the embedding model's training set (ImageNet1K training). We compare the magnitude of the corresponding predictive variances with the extent to which the data is OOD relative to the training set. We consider CelebA \cite{liu2015faceattributes}, a mammography dataset and lightly augmented versions of ImageNet1K (see Fig. \ref{fig:augs}), to observe whether the magnitude in the predictive variance magnitude correlates with the \textit{assumed} effectiveness of the FID (i.e. worse performance for datasets that are more OOD relative to the training data). This assumption is drawn from other work that has discussed how the FID is known to be less effective when used on non-natural and OOD datasets \cite{konz2024fr,wu2025pragmatic,truong2021transferable,kynkaanniemi2022role} (aside from a single work with limited scope suggesting otherwise \cite{woodland2024feature}). The characteristic feature representations of $D_t$ are generally less likely to be found in $D_s$ or to be effectively extracted/encoded by $G_s$ the more OOD they are relative to $D_s$. However, we do not validate this assumption here, as it requires a metric for the effectiveness of the FID, which is challenging to formulate. This comparison akin to an uncertainty reliability analysis, where uncertainty magnitude is compared to the magnitude of prediction error \cite{guo2017calibration} (though here, we do not formally justify whether the predictive variance here can be considered an uncertainty). 

We also consider several other noise augmented versions of ImageNet1K's validation set (additive random Gaussian noise with a standard deviation defined by X~$\%$ of the image's maximum amplitude, where $X$ is the `strength' of the augmentation) to measure the sensitivity of the predictive variance to gradual changes in image content. We coarsely validate whether the direction of change in the FID are sensible by comparing them to changes in mean image-wise mean absolute error (MAE), the multi-scale structural similarity index (MS-SSIM) \cite{wang2003multiscale} between the original and augmented images (model-free metrics for content similarity, that encode information about structural preservation), and the top-5 accuracy on ImageNet1K classification task. 

Given we do not use a direct measure of the effectiveness in the FID, we can not make strong conclusions about the how well our MCD-derived predictive variances indicate its trustworthiness. Nonetheless, these experiments provide useful information about how the predictive variance changes with respect to varying properties of the data, and give some insight into strategies that can be used to evaluate the practical utility of this approach that may be improved or used more effectively in future work. 

\subsection{Related work}
There is some precedent for using uncertainty quantification techniques to detect OOD inputs and to evaluate the effectiveness of a feature embedding model's ability to encode characteristic feature representations. \cite{nguyenstochastic} applied MCD to the penultimate layer of a classifier model, and described how OOD examples had correspondingly lower predictive variance in the model's embedding representations. The authors hypothesised that the higher norms of the embedding representations of independent and identically distributed (IID) examples that emerge as a consequence of learning a classification task (neural collapse) are responsible for this behaviour. Using this principle, they developed their own method for OOD detection, but did not consider how this could be used to assess the trustworthiness of synthetic image quality metrics that utilise embedding representations like the FID. 

In other work \cite{bench2025quantifying} MCD was applied to an autoencoder model trained on a subset of the Imagenette dataset (subset of ImageNet1K), where the uncertainty in the bottleneck representation was used to provide a heuristic measure of the trustworthiness of the Fr\'{e}chet Autoencoder Distance (FAED) - the autoencoder analogue of the FID \cite{buzuti2023frechet}. Here, both expressions of predictive variance considered were mostly found to \textit{increase} with the extent to which the input data was OOD of the training data. However, this correlation was not observed for the variance in the embedding representations, where some forms of data augmentation applied to Imagenette examples (additive noise) resulted in a \textit{decrease} in the predictive variance. The insights described in \cite{nguyenstochastic} suggest that the embedding representations of a classifier model may have unique characteristics relative to those learned from optimising other objectives/architectures; this possibly explains the different behaviours observed in the embeddings under stochastic sampling. Furthermore, the scale and diversity of the datasets differ considerably between the two studies, which may also have an effect on the properties/behaviour of the embeddings. The scale of ImageNet1K may also influence the behaviours of embeddings learned by the InceptionV3 used by the FID. Therefore, these works alone do not provide a complete picture of how UQ techniques may provide an assessment of the trustworthiness of the FID, and motivates the study described in this work.

\section{Methods}

We train an InceptionV3 architecture on ImageNet1K with dropout regularisation applied to every convolutional layer throughout the architecture using the same implementation details as the InceptionV3 paper \cite{szegedy2016rethinking} initialised with pretrained weights from training on ImageNet1K without dropout. The model was trained with a batch size of 32 for 91 epochs (early stopping), and a learning rate of 0.01.

Once trained, each test batch is evaluated J=20 times (the maximum number of samples allowed, given restrictions on storage). We then acquire the average of the normalised trace of the covariance in the test embeddings, denoted as $\text{pVar}$, given embeddings $l_{i,j,k}$, where $i$ indexes the input image, and $k$ is each element of the embedding ($K=2048$ elements, and $j = 1,2,...,J$),

\begin{equation}
\label{eq:pvar}
    \text{pVar} = \frac{1}{I}\sum^{I}_{i=1}\frac{1}{K(J-1)}\sum^{J}_{j=1}||\textbf{l}_{i,j} - \bar{\textbf{l}}_i||^2_2    
\end{equation}
where $\bar{\textbf{l}}_i = \frac{1}{J}\sum^{J}_j \textbf{l}_{i,j}$. As discussed in \cite{nguyenstochastic} this may provide some proxy indication of whether the inputs are outliers with respect to the training set, which may relate to whether they encode characteristic feature representations of the images and hence the trustworthiness of the FID. Other work has discussed the association of embedding stochasticity with uncertainty \cite{an2023maximum,ahn2024uncertainty,scott2019stochastic,oh2018modeling}, which may also reflect how well the model encodes characteristic feature representations in the embedding representations. We use this expression of predictive variance to supplement discussion of the results, and evaluate its use as a possible proxy metric for the trustworthiness in the FID.

We also compute the FID for each evaluation $j$ of the test set $\hat{X}$ with a fixed reference set of images $X$,
\begin{equation}
\label{eq:fd}
        \mathrm{FID}^{j}(\hat{X},X) = \\\left\|\mu^{j}_{\hat{x}}-\mu_x\right\|_2^2+\text{tr}\left(\Sigma^{j}_{\hat{x}}+\Sigma_x-2\left(\Sigma^{j}_{\hat{x}} \Sigma_x\right)^{\frac{1}{2}}\right),
\end{equation}
where $\mathcal{N}_{{\hat{x}}}(\mu^{j}_{\hat{x}}, \Sigma^{j}_{\hat{x}})$ is a multivariate Gaussian parameterised by the mean of the embeddings from $\hat{X}$, $\mu^{j}_{\hat{x}}$, and their covariance $\Sigma^{j}_{\hat{x}}$, where $\mathcal{N}_{x}\left(\mu_x, \Sigma_x\right)$ is the corresponding distribution for the reference set. The variance of the resultant FIDs is computed,
\begin{equation}
\label{eq:vfd}
\begin{split}
        \text{vFID}(\hat{X},X) = \left[\sigma_{j} \left( \mathrm{FID}^{j}(\hat{X},X) \right)\right]^2,
    \end{split}
\end{equation}
where $\sigma_{j}$ takes the standard deviation of the $J$ different values of FID$^{j}$.

Here the reference set of embeddings is acquired by feeding one half of ImageNet1K's validation set into the model without dropout active unless otherwise specified. vFID (or the corresponding standard deviation $\sigma$FID) is used as our heuristic expression of the trustworthiness of the FID.

\label{sec:sim_sem}

\subsection{Uncertainty quantification}
\subsubsection{Equal augmentation}
We assess how the predicted variance in the FID changes as image contents in the test and reference sets become more homogeneous and similar as a result of noise augmentation (additive Gaussian noise, where the strength is given by  the \% of the max value of the image used to parameterise the standard deviation). This is achieved by adding progressively larger amounts of noise \textit{equally} to two halves of Imagenet1K. Any increase in the FID would indicate that the metric is not effective, and  correspondingly, reliable expressions for the trustworthiness of the FID (e.g. $\sigma$FID) should increase in magnitude. In contrast, a consistent decrease in the FID with augmentation strength, trending to 0, would indicate that the metric is effective to some degree. In this case reliable uncertainties should be low in magnitude. Here, the FID is computed using a reference set of latents output by the embedding model with dropout active. We choose noise augmentation instead of other approaches given it will homogenise contents in the limit of stronger augmentations, in contrast to overlaid image augmentation.
\begin{figure}[h!]
    \centering
    \includegraphics[width=.95\linewidth]{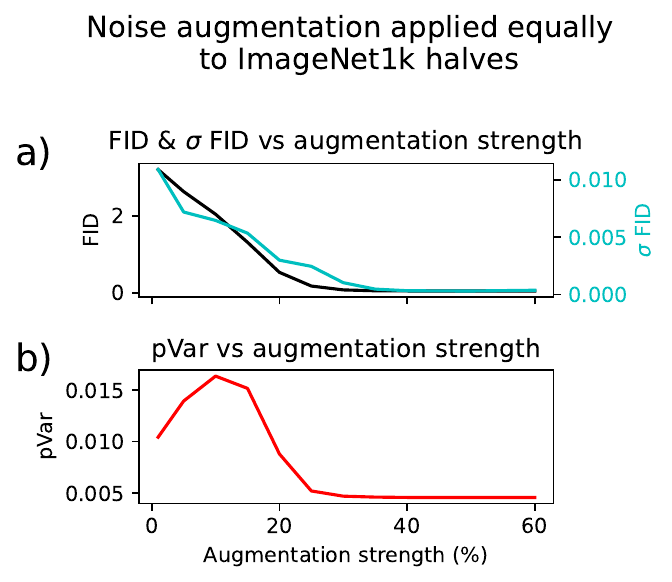}
    \caption{a) FID and $\sigma$FID vs. augmentation strength and b) pVar vs augmentation strength for the equal augmentation experiment.}
    \label{fig:equal_aug}
\end{figure}
\subsubsection{OOD datasets}
FID is less dependable when used with datasets that are non-natural or OOD. In the absence of other quantitative means to assess the effectiveness of the FID, we analyse how the magnitude of the predictive variance changes as the test data becomes increasingly OOD relative to the training set as a coarse means to assess whether pVar and $\sigma$FID can be used to assess the effectiveness of the FID. I.e. if one can assume that the effectiveness of the FID decreases for input data that is more OOD, then pVar and $\sigma$FID should correlate in magnitude if they are suitable indicators of the trustworthiness of the FID. We consider CelebA and a set of mammography images as our external datasets. We use the mean k-NN distance (k=5) on the L2 normalised latent representations fit to the normalised ImageNet1K reference latents to verify the extent to which image sets are OOD relative to ImageNet1K. We also consider lightly augmented variants of ImageNet1K: one version augmented with four random Imagenet1K images, a version augmented with four random CelebA images, and a version augmented with four random mammography images. We supplement validation of k-NN by providing the top-5 accuracy for ImageNet1K-based datasets, which should decrease as k-NN increases if k-NN is a reliable indicator of the extent to which a dataset is OOD. It also provides an indirect means to evaluate the effectiveness of the FID (akin to \cite{wu2025pragmatic}, where higher FIDs should correspond to higher prediction error) which in principle can supplement discussion of how well the predictive variance indicates the trustworthiness of the FID.

\subsection{Noise augmentation sensitivity analysis}
We also consider other noise augmented variants of the ImageNet1K validation set. We tune the strength of the noise augmentation to assess the sensitivity of pVar and $\sigma$FID to gradual changes in image features. To provide some coarse indication of the trustworthiness of the FID, we compare these changes in the FID with changes in the MAE, and MS-SSIM between the original and augmented images, as they are model-free metrics for content similarity (though just focus on structural similarity). As with the unaugmented external datasets, we also consider the top-5 accuracies on the ImageNet1K classification task. An effective FID should have a significant negative correlation with top-5 accuracy as augmentation strength increases. Positive correlation between top-5 accuracy and FID should correspond to low trustworthiness in the FID according to the proxy measures of its trustworthiness (i.e. $\sigma$FID and pVar) if they are reliable. While this validation methodology lacks quantitative rigour needed to make strong conclusions about how well $\sigma$FID and pVar indicate the trustworthiness of the FID (a metric for the effectiveness of the FID should quantify something about how well characteristic feature representations are encoded in the embeddings), the use of MAE, MS-SSIM, and top-5 accuracy nonetheless provide some notion of the effectiveness of the FID that enables some assessment of our metrics.

\section{Results and discussion}

\begin{figure*}[h!]
    \centering
    \includegraphics[width=.9\columnwidth]{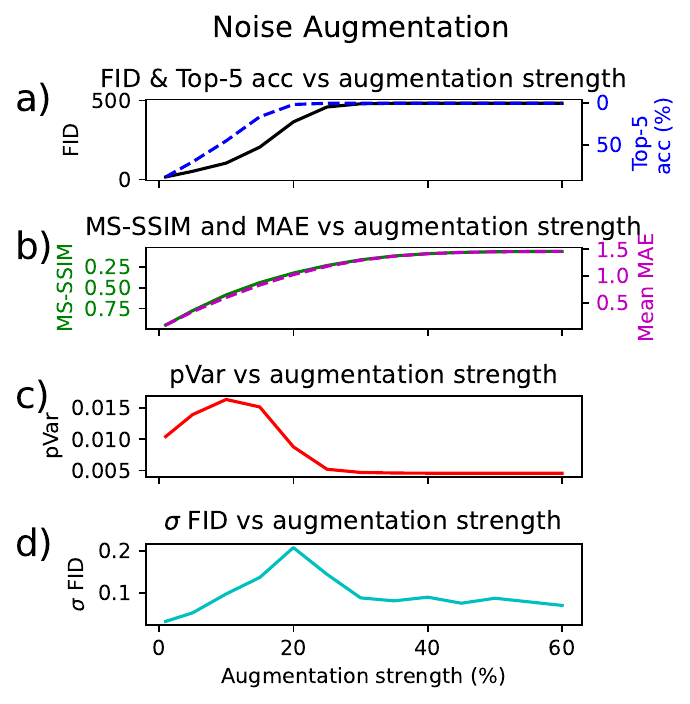}
    \includegraphics[width=.8\columnwidth]{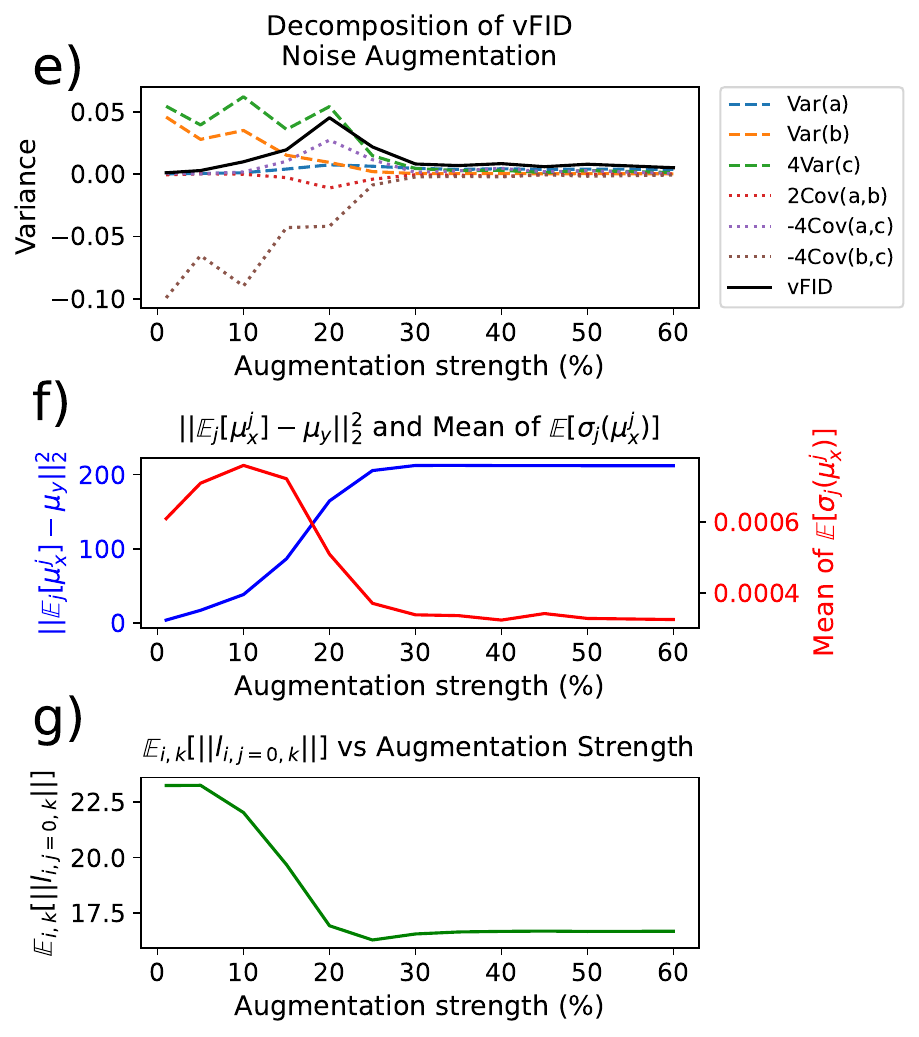}
    \includegraphics[width=1.9\columnwidth]{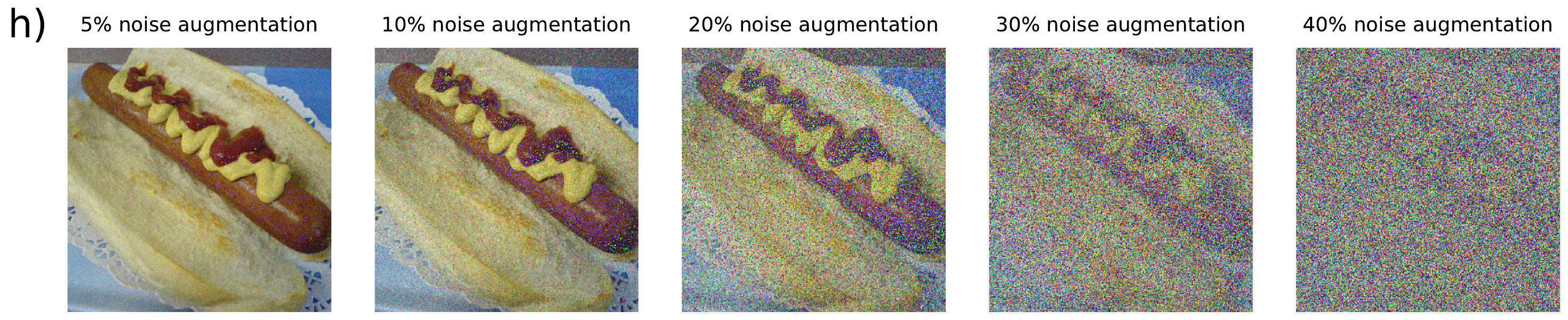}
    
    \caption{Plots of a) Mean FID over MCD samples, and (inverted vertical axis) top-5 accuracy on ImageNet1K classification, b) MAE and (inverted vertical axis) MS-SSIM, c) pVar, and d) $\sigma \text{FID}$ vs noise augmentation (additive random Gaussian noise with a standard deviation defined by some percentage of the image's maximum amplitude, referred to here as strength). e) Decomposition of vFID vs augmentation strength where $a = \|\mu^{j}_{\hat{x}}-\mu_y\|_2^2$, $b = \text{tr}(\Sigma^{j}_{\hat{x}}+\Sigma_y)$ and $c = \text{tr}(\Sigma^{j}_{\hat{x}} \Sigma_y)^{\frac{1}{2}}$, so $\text{FID} = a + b - 2c$. f) First term of the FID computed using the mean over the MCD samples of the test latent means (blue) and expected standard deviation of the latent means across MCD samples (red) vs augmentation strength; the latter correlates with c) as expected. g) Norm of the test embeddings vs augmentation strength, h) examples of noise augmented images.}
    \label{fig:sens_inputs}
\end{figure*}

\begin{table}[h!]
\centering
\caption{FID, pVar, and $\sigma$FID for various test datasets. k-NN score indicates the extent to which data is OOD relative to the reference set of ImageNet1K (higher corresponds to more OOD).}
\begin{tabular}{|l|c|c|c|c|c|}
\hline
\textbf{Dataset} & \textbf{FID} & \textbf{$\sigma$FID } & \textbf{pVar} &\textbf{ k-NN} & \textbf{top-5 acc} \\
\hline
ImageNet1K (I1k) & 8 & 0.009 & 0.028 & 0.61&93 \%\\
\hline
I1k + 1\% noise & 16  & 0.035  & 0.010 & 0.63&88 \%\\
\hline
I1k + 4 CelebA & 62 & 0.056 & 0.015 & 0.69&70 \%\\
\hline
I1k + 4 Mammo  & 76 & 0.100 & 0.014 & 0.70&67 \%\\
\hline
CelebA & 321  & 0.110  & 0.005  & 0.75&-\\
\hline
Mammography & 365 & 0.350 & 0.011 & 0.80&-\\
\hline
\end{tabular}
\label{tab:res}
\end{table}

\subsection{Equal augmentation}
Fig. \ref{fig:equal_aug} shows that the FID decreases with increasing augmentation strength applied to both halves of ImageNet1K. This coarsely validates its effectiveness on noise augmented data when both the test and reference sets are augmented, as we would expect the metric to decrease and trend to zero as contents homogenise. Correspondingly, we find that $\sigma$FID decreases, and critically, exhibits low magnitude relative to the sensitivity study shown in Fig. \ref{fig:sens_inputs}d where image contents become less similar. Given the FID appears accurate and is low magnitude, this suggests that $\sigma$FID could be a reliable indicator of the effectiveness of the FID.

\subsection{OOD datasets and sensitivity analysis}
Table \ref{tab:res} shows how $\sigma$FID increases with the extent to which the test data was OOD of the training data. If one can assume that the effectiveness of the FID for data that is increasingly OOD will decrease, then this suggests $\sigma$FID may indicate the trustworthiness of the FID to some extent. Though, the validity of this assumption needs to be verified to draw stronger conclusions.
pVar on the other hand does not exhibit a consistent trend, which could suggest poor suitability as an indicator for OOD detection, or as a proxy measure of the effectiveness of the FID. The larger values for overlay augmentations (suggesting higher trustworthiness and that the data is less OOD \cite{nguyenstochastic}) relative to noise augmentation may be due to the fact that much of the original images are unchanged as only local regions of the image are affected with the overlay augmentations.

The sensitivity study in Fig. \ref{fig:sens_inputs} shows that both $\sigma$FID and pVar initially increase and then decrease with augmentation strength. pVar may decrease at higher augmentation strengths due to the arguments outlined in \cite{nguyenstochastic}; indeed we find that the norm of the embeddings decreases with higher augmentation strengths (Fig. \ref{fig:sens_inputs}g) where the decrease in pVar corresponds to the decrease in the mean norm of the latents. 

An initial increase in $\sigma$FID was not observed for the equal augmentation case (see Fig. \ref{fig:equal_aug}) despite the same base sets of images being used. This result suggests that \textit{differences} in the fidelity of image features/content between reference and test sets (along with the features themselves) can significantly affect uncertainty magnitude for OOD data. I.e., the increase in $\sigma$FID observed here is because of the more significant difference in the properties between the sets of images compared to the equal augmentation case despite the extent of augmentation being similar. At extreme strengths, the lack of structure/homogenised representation corresponds to less variable embeddings and lower $\sigma$FID. 

The behaviour of the $\sigma$FID can be further investigated by observing how the variance of each term behaves with increasing augmentation strength. To simplify notation, we define $a = \|\mu^{j}_{\hat{x}}-\mu_y\|_2^2$, $b = \text{tr}(\Sigma^{j}_{\hat{x}}+\Sigma_y)$ and $c = \text{tr}(\Sigma^{j}_{\hat{x}} \Sigma_y)^{\frac{1}{2}}$, so $\text{FID} = a + b - 2c$, where $\mu_y$ and $\sum_y$ denote the same quantities as $\mu_x$ and $\sum_x$, respectively. The decomposition of the variance is given by:
\begin{equation}
\begin{aligned}
    \text{vFID} &= \operatorname{Var}(a) + \operatorname{Var}(b) + 4\operatorname{Var}(c) \\
    &\quad + 2\operatorname{Cov}(a,b) - 4\operatorname{Cov}(a,c) - 4\operatorname{Cov}(b,c).
\end{aligned}
\end{equation}

Each term is plotted in Fig. \ref{fig:sens_inputs}e for noise augmentation. Here, we see that all terms gradually trend to zero with increasing augmentation strength. The changes in each term depend on the interplay of several factors. E.g. for term $a$, the differences in the means increase with augmentation strength (blue line in Fig. \ref{fig:sens_inputs}f), so given its quadratic form, the overall variance of the term increases despite the variance in test mean term decreasing at larger augmentation strengths (red line in Fig. \ref{fig:sens_inputs}f). At very high augmentation strengths, the variation in the test mean is so low (red line in Fig. \ref{fig:sens_inputs}f), that even with a considerable difference between the two means, the variance of the whole term decreases.

\begin{figure}
    \centering
    \includegraphics[width=0.99\linewidth]{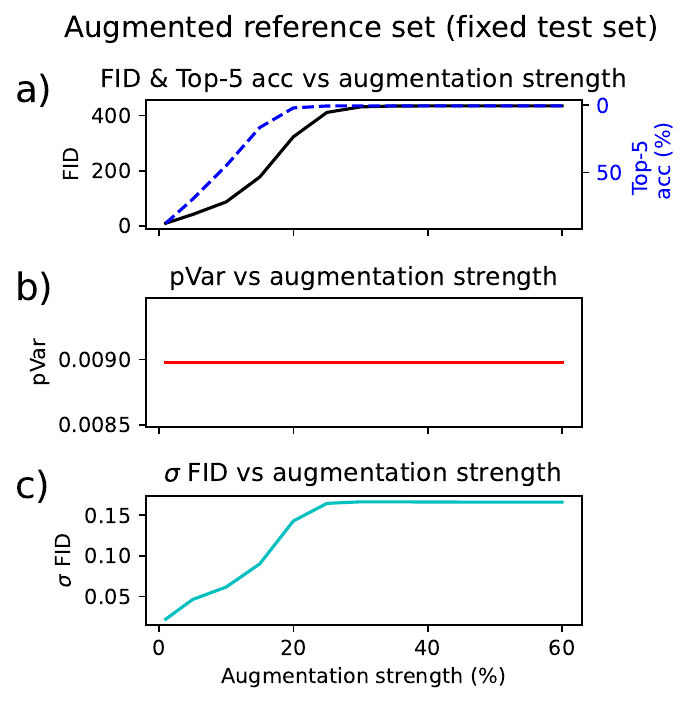}
    \caption{Results of the same experiment as Fig. \ref{fig:sens_inputs} but with a fixed test set (MCD applied to unaugmented inputs, keeping pVar constant for the plot) and increasingly augmented reference set (single MCD sample). Here, we do not see a drop in $\sigma$FID in plot c, which suggests the discrepancy in image fidelity could be the reason for the increase observed in Fig. \ref{fig:sens_inputs}d, and that the decrease at higher strength augmentations is due to the decrease in pVar.}
    \label{fig:fixed_set}
\end{figure}

It is worth noting that the decrease in $\sigma$FID in the sensitivity experiment (Fig. \ref{fig:sens_inputs}) corresponds to larger augmentation strengths that qualitatively result in the destructuring and homogenisation of image contents (see top-5 acc. plateau). The cause of this behaviour is unclear given the complex interplay of the various terms of the FID. But, it suggests that complexity of feature representations is a significant factor determining the magnitude of $\sigma$FID. To isolate the role of pVar (i.e. keep this factor constant), Fig. \ref{fig:fixed_set} shows the results of the same experiment as Fig. \ref{fig:sens_inputs}, but with a fixed test set, and an increasingly augmented reference set. Here, we find $\sigma$FID increases and then plateaus as the reference set becomes increasingly augmented. This suggests that, akin to term $a$ in the variance decomposition, $\sigma$FID scales with the extent of the difference between the test and reference sets, as well as the magnitude of pVar.   

The existing discussion of the effectiveness of the FID on OOD datasets in the literature focus on non-natural images that have not been heavily augmented. So our reliance on the assumed effectiveness of the FID on OOD data as a method for coarsely validating the suitability of $\sigma$FID and pVar as trustworthiness metrics doesn't necessarily apply to heavily augmented data, limiting our ability to make observations about their suitability for these higher strengths.

\section{Conclusion}

Our results indicate that both pVar and $\sigma$FID depend strongly on the properties of the input data. $\sigma$FID generally correlates with the extent to which unaugmented datasets are OOD of ImageNet1K, and therefore, likely provides some indication for when the FID may be less informative/effective. pVar did not exhibit the same correlation, which suggests it is less reliable as a proxy measure a) of the extent to which data is OOD (despite being used as such in the literature \cite{nguyenstochastic}) and b) the effectiveness of the FID. We find that the magnitude of $\sigma$FID in the equal augmentation experiment decreases with the FID as augmentation strength increases, and generally has lower magnitude than the equivalent result from the sensitivity study, which may also suggest some degree of reliability as a proxy for the trustworthiness of the FID. 

The sensitivity study using noise augmented data showed that the $\sigma$FID exhibited an increase with augmentation strength at lower strengths, but then decreased at higher strengths corresponding to the destructuring and homogenisation of image contents. pVar exhibited similar trends, where the decrease at high augmentation strengths is in line with the hypothesis proposed in \cite{nguyenstochastic}. The cause of the decrease in $\sigma$FID at higher strengths requires a closer analysis of the complex interplay of the various terms of the FID (similar analysis is needed for explaining the initial increase in pVar at low augmentation strengths). The differences in the behaviour of $\sigma$FID in the equal augmentation and sensitivity experiments (both Figs. \ref{fig:sens_inputs} and \ref{fig:fixed_set}) demonstrate that the magnitude of $\sigma$FID scales with the difference in image content between test and reference sets, and pVar. 

If the assumption that the FID performs worse on OOD data is valid, then the results in Table \ref{tab:res} for the OOD datasets could suggest that $\sigma$FID exhibits some degree of reliability. Following this, the highest magnitude $\sigma$FID values were found for the mammographic image set (the most OOD), which may suggest that the metric is the least trustworthy for these images. But it is evident that a quantitative `gold standard' metric for assessing the effectiveness of the FID is needed to robustly validate the suitability of the expressions of the trustworthiness of the FID presented here, and hence, the practical utility of the stochastic embedding approach (in particular for the sensitivity study involving augmented data). Future work could consider more careful choice of augmentation strategy/datasets, and downstream tasks to enable a more effective assessment of the effectiveness of the FID (i.e. impose some error as observed in augmentation studies in \cite{deo2025metrics}) and therefore reliability of $\sigma$FID using our proposed experiments. 

Nonetheless, our results provide useful insight into how our heuristic metrics for the trustworthiness of the FID change with respect to varying datasets, giving some indication of how uncertainty quantification techniques can be used to evaluate the trustworthiness of the FID. This work establishes a framework for uncertainty-aware evaluation of generative models, enabling trust assessment in high-stakes applications such as healthcare and other safety-critical domains.

\bibliographystyle{unsrt}
\bibliography{bib}

\end{document}